\documentclass[10pt,twocolumn,letterpaper]{article}

\usepackage{iccv}
\usepackage{times}
\usepackage{epsfig}
\usepackage{graphicx}
\usepackage{amsmath}
\usepackage{amssymb}
\usepackage{booktabs}
\usepackage{caption}
\usepackage{float}
\usepackage{multirow, multicol}
\usepackage[table]{xcolor}

\usepackage[breaklinks=true,bookmarks=false]{hyperref}

\iccvfinalcopy 

\ificcvfinal\fi

\begin{document}

\title{SAM3D: Zero-Shot 3D Object Detection via Segment Anything Model}

\author{Dingyuan Zhang$^1$, Dingkang Liang$^1$, Hongcheng Yang$^1$,\\ Zhikang Zou$^2$, Xiaoqing Ye$^2$, Zhe Liu$^1$,  Xiang Bai$^{1\dag}$\\
$^1$Huazhong University of Science and Technology, \{dyzhang233, dkliang, xbai\}@hust.edu.cn\\
$^2$Baidu Inc., China\\
{\tt\small \url{https://github.com/DYZhang09/SAM3D}}
}

\maketitle
\protect \renewcommand{\thefootnote}{\fnsymbol{footnote}}
\footnotetext{$^{\dag}$Corresponding author.} 
\ificcvfinal\thispagestyle{empty}\fi


\section{Introduction}
In the past few years, foundation models have thrived and succeeded in linguistic and visual tasks, showing astonishing zero-shot and few-shot capabilities. Their advances encourage researchers and industries to extend the boundaries of what artificial intelligence can do and have shown some fantastic products (e.g., ChatGPT~\cite{brown2020gpt3}) with the potential to change the world.

Recently, Kirillov et al.~\cite{kirillov2023sam} proposed a new vision foundation model for image segmentation, the Segment Anything Model (SAM), trained on a huge dataset called SA-1B. The flexible prompting support, ambiguity awareness, and vast training data endow the SAM with powerful generalization, enabling the ability to solve downstream segmentation problems using prompt engineering. Some following works leverage the excellent zero-shot capability of SAM to solve other 2D vision tasks (e.g., medical image processing~\cite{zhou2023samforpolyps} and camouflaged object segmentation~\cite{tang2023sammeetscamouflaged}). Although SAM presents great power on some 2D vision tasks, whether it can be adapted to 3D vision tasks still needs to be discovered. With this inspiration, a few works attempt to combine SAM with pre-trained 3D models to learn 3D scene representation (e.g., SA3D~\cite{cen2023sa3d}) and single-view reconstruction (e.g., anything-3D~\cite{shen2023anything3d}), showing promising results.

3D object detection, one of the fundamental tasks in 3D vision, has a wide range of real-world applications (e.g., autonomous driving). Although plenty of works aim to solve this task, the zero-shot setting on 3D object detection still needs to be explored. Thus, considering the advance of SAM, it is natural to question: \textit{Can we adapt the zero-shot capability of SAM to 3D object detection}? 

In this paper, we aim to explore the zero-shot 3D object detection with SAM~\cite{kirillov2023sam} alone. Considering SAM is initially built for 2D images, many challenges exist when using SAM for 3D detection (Please refer to the appendix for more discussion). The key insight is that we can leverage the powerful capability of SAM for 3D object detection by using the Bird's Eye View (BEV), which carries crucial 3D information (e.g., depths) with a 2D image-like data format. Thus, the challenges to using SAM for 3D detection can be significantly solved. With this observation, we present SAM3D, which uses SAM to segment on BEV maps and predicts objects based on the masks from its outputs.

We evaluate our method on the large-scale Waymo Open Dataset~\cite{sun2020waymo}, and the results show the great potential of SAM on 3D object detection. Although this paper is only an early attempt, it gives a positive signal for applying vision foundation models like SAM for 3D vision tasks, especially for 3D object detection.

\begin{figure*}[t]
\begin{center}
\includegraphics[width=0.9\linewidth]{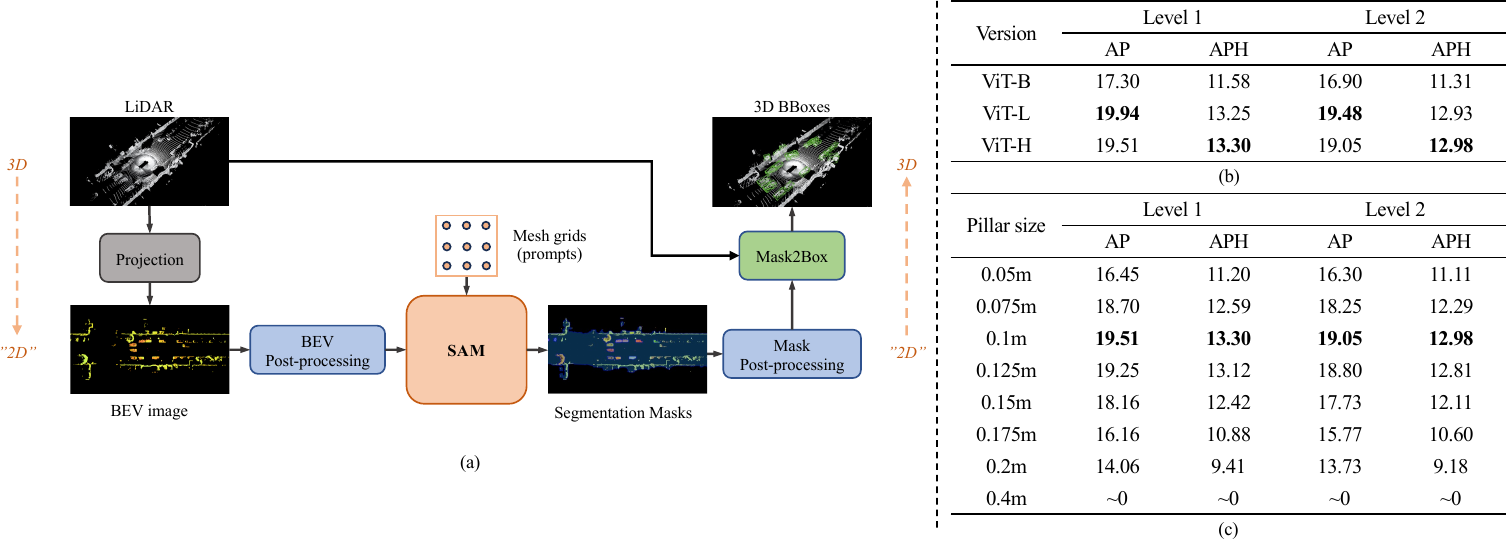}
\end{center}
\caption{(a) The overall framework of our method. We first project LiDAR points to colorful BEV images via a predefined palette, then post-process BEV images to better fit the requirements of SAM. After the segmentation, we post-process the noisy masks and finally predict 3D bounding boxes with the aid of LiDAR points. (b) The results of SAM3D using different versions of SAM. (c) The results of SAM3D using different pillar sizes. We report metrics of VEHICLE in the range [0,30) on Waymo \textit{validation} set.}
\label{fig:pipeline}
\end{figure*}

\section{Method}
We consider point cloud as the input of our method, which is a 3D representation and naturally sparse, while SAM is trained for 2D images with dense semantics. Our basic idea is to translate LiDAR points into a 2D image-like representation with 3D information that narrows the domain gap, thus BEV is a straightforward choice. We build the whole pipeline with SAM based on BEV, shown as Fig.~\ref{fig:pipeline}(a).
Our method mainly contains five steps:

Firstly, our method conducts the LiDAR-to-BEV projection, which translates sparse LiDAR signals to discriminative BEV images. At this step, we use the projection equations to determine each point's coordinate on the image plane and a predefined intensity-to-RGB mapping to get RGB vectors for pixels in a BEV image, making it more discriminative during processing.
    
Then, the BEV post-processing modifies the original BEV images with the morphology dilation (interpreted as a max pooling) since SAM is trained on natural images with ``dense" signals, which differs from the ``sparse" BEV images. This step helps form more suitable inputs for SAM, leading to easier segmentation and better performance. 

After obtaining the desired BEV images, we segment the BEV images using SAM, which supports various prompts like point, box, and mask prompts. Our goal in this step is to segment foreground objects as many as possible, so we choose to cover the whole image with mesh grid prompts. Additionally, we prune the prompts in this step without performance sacrifice to accelerate the segmentation.

Despite SAM's powerful zero-shot capability, a non-negligible domain gap still exists. Hence, we propose mask post-processing for filtering noisy masks according to some rules drawn from priors, which reduces the number of false positives and helps improve the final performance.
    
Finally, after the segmentation and post-processing, we predict 3D bounding boxes from the foreground masks. Since BEV images already carry depth information, we can directly estimate the horizontal attributes (i.e., horizontal object center, length, width, and heading) of 3D bounding boxes from the 2D masks. Meanwhile, for the vertical attributes (i.e., vertical object center and height), LiDAR points will be utilized as extra information compensation. 

Please refer to the appendix for more detailed methods.

\section{Experiments}
We evaluate our method on the Waymo Open Dataset~\cite{sun2020waymo}, one of the large-scale datasets for autonomous driving. The dataset is split into 798 training sequences, 202 validation sequences, and 150 testing sequences. Since our method performs zero-shot object detection, we only focus on the validation sequences. For the metrics, because of the natural sparsity of point clouds and the lack of semantic label outputs, we only care about the mAP and mAPH of VEHICLE with a distance of at most 30 meters in this paper.

Since SAM uses different backbones with different complexities, we conduct experiments to evaluate the effectiveness of our method, shown in Fig.~\ref{fig:pipeline}(b). It reveals that using SAM with less capacity performs worse. However, there is only a marginal difference between SAM with ViT-L and ViT-H. We argue that the model capacity is not the performance bottleneck when using large models, and the power of SAM still needs to be fully unleashed. For insurance purposes, we use SAM with ViT-H. 
We also conduct experiments to determine how the pillar size influences the performance in Fig.~\ref{fig:pipeline}(c). When using larger pillar sizes such as $0.2m$ and $0.4m$, the discretization errors are relatively large, and it is hard to distinguish different objects when they are close to each other. However, pillar sizes that are too small also harm performance. One possible reason is that due to the high resolution of the small pillar size and the sparsity of LiDAR signals, it is difficult for individual instances to form a completely connected region. SAM tends to separate one object into many parts. We set the pillar size as $0.1m$, which is a good balance. Please refer to the appendix for all detailed results.

\section{Conclusion}
This paper explores the zero-shot 3D object detection with the visual foundation model SAM and proposes the SAM3D. To narrow the gap between the training data of SAM and 3D LiDAR signals, we use the BEV images to represent 3D outdoor scenes. We propose a SAM-powered BEV processing pipeline to utilize the great zero-shot capability of SAM for zero-shot 3D object detection. Qualitative and ablation experiments on the Waymo open dataset show promising results for adapting the zero-shot ability of SAM to 3D object detection. Although this paper is only an early attempt, we believe it presents a possibility and opportunity to unleash the power of foundation models like SAM on 3D tasks with technologies like few-shot learning, model distillation, and prompt engineering in the future. The code has been released in \url{https://github.com/DYZhang09/SAM3D}.

\section{Acknowledgements}
This work was supported in part by the National Science Fund for Distinguished Young Scholars of China (Grant No. 62225603), in part by the Hubei Key R\&D Program (Grant No. 2022BAA078), and part by the "Qisun Ye" Science Fund (U2341227).

\section{Appendix}
\subsection{Related Work}
\subsubsection{2D tasks with SAM}
Kirillov et al.~\cite{kirillov2023sam} proposed a new vision foundation model for image segmentation, the Segment Anything Model (SAM), trained on a huge dataset called SA-1B to solve the segment anything task. The flexible prompting support, ambiguity awareness, and vast training data endow the SAM with powerful generalization, enabling the ability to solve downstream segmentation problems using prompt engineering, inspiring many following works. SAMPolyp~\cite{zhou2023samforpolyps} applies SAM to the polyp segmentation task under the unprompted settings. Deng et al.~\cite{deng2023samforpathology} assess the zero-shot segmentation performance of the SAM model on digital pathology tasks and find that SAM achieves remarkable segmentation performance for large connected objects. He et al.~\cite{he2023samfor12medicaldatasets} test SAM’s accuracy in 12 medical image datasets, revealing that SAM is more accurate in 2D medical images, larger target region sizes, and easier cases. SAMCOD~\cite{tang2023sammeetscamouflaged} evaluates SAM’s performance on camouflaged object detection (COD) benchmarks, indicating that SAM can achieve a noteworthy performance compared to some COD-oriented methods. Ji et al.~\cite{ji2023samstrugglesinconcealed} compare SAM with cutting-edge methods and observe the limited power of SAM in concealed scenes. These works focus on utilizing the SAM on 2D tasks like medical image analysis and camouflaged object detection, while our method explores the ability of SAM on the 3D perception task.

\subsubsection{3D tasks with SAM}
Although SAM presents great power on some 2D vision tasks, whether it can be adapted to 3D vision tasks still needs to be discovered. With this inspiration, a few works attempt to combine SAM with pre-trained 3D models. SA3D~\cite{cen2023sa3d} generalizes SAM to segment 3D objects by leveraging the Neural Radiance Field (NeRF) as a cheap and off-the-shelf prior, which constructs the 3D mask of the target object via alternately performing the proposed mask inverse rendering and cross-view self-prompting across various views. Anything-3D~\cite{shen2023anything3d} combines BLIP, a pre-trained 2D text-to-image diffusion model, and SAM for the single-view conditioned 3D reconstruction task, showing promising results. 3D-Box-Segment-Anything~\cite{chen20233D-Box-Segment-Anything} utilizes SAM with a pre-trained 3D detector VoxelNeXt~\cite{chen2023voxelnext} for interactive 3D detection and labeling. Unlike them, our method explores zero-shot 3D object detection with SAM alone (i.e., it DOES NOT rely on any other pre-trained model).

\subsubsection{3D object detection}
3D object detection, one of the fundamental tasks in 3D vision, has a wide range of real-world applications (e.g., autonomous driving). There are plenty of works~\cite{yan2018second, lang2019pointpillars, shi2020pvrcnn, chen2023voxelnext, li2023dds3d, zhang2023vitwss3d, zhou2023diffusion, xiong2023you} aim to solve this task. SECOND~\cite{yan2018second} and PointPillars~\cite{lang2019pointpillars} convert the point cloud into grid-based representation and introduce the sparse convolution and pillars representation, respectively, to solve the 3D object detection with low time consumption. PointRCNN~\cite{shi2019pointrcnn} proposes to process the point cloud directly and segment the point cloud before the second-stage refinement for generating high-quality proposals. PVRCNN~\cite{shi2020pvrcnn} combines 3D CNN and point-based operations to learn more representative features. VoxelNeXt~\cite{chen2023voxelnext} introduces a fully sparse voxel-based pipeline for 3D object detection without relying on hand-crafted proxies, achieving a better speed-accuracy trade-off. QTNet~\cite{hou2023qtnet} proposes a new query-based temporal fusion method to facilitate temporal detection efficiency. ViT-WSS3D~\cite{zhang2023vitwss3d} offers a novel weakly semi-supervised paradigm to lower the dependencies on expensive 3D annotations. Although 3D object detection with 3D annotations has been widely studied, the zero-shot setting on 3D object detection, which has significant practical value (e.g., low-cost data labeling), still needs to be explored. In this paper, we aim to explore zero-shot 3D object detection with the aid of SAM.

\subsection{Preliminary}
\subsubsection{Definitions of 3D object detection}
\label{appendix:def_of_3DDet}
Similar to 2D object detection, the goal of 3D object detection is to predict the locations and categories of all objects of interest given the perceptive sensor data (e.g., LiDAR points for LiDAR-based 3D object detection and camera images for camera-based 3D object detection). In this paper, we focus on the LiDAR-based 3D object detection. To be more specific, given LiDAR points $P = \{ (x_i, y_i, z_i) \}_{i=1}^{N_{p}}$ of the scene, 3D detectors need to infer about all objects $O = \{ (c_i, B_i^{3D}) \}_{i=1}^{N_o}$ in the scene, where $c_i$ and $B_i^{3D}$ are the category and geometric attributes of the $i$-th object, and $N_o$ is the number of objects. Typically, we define the geometric attributes as object centers, dimensions, and orientations, formally written as:
\begin{equation}
    B_i^{3D} = (x^{3D}_i, y^{3D}_i, z^{3D}_i, dx^{3D}_i, dy^{3D}_i, dz^{3D}_i, \theta^{3D}_i),
\end{equation}
where $(x^{3D}_i, y^{3D}_i, z^{3D}_i)$, $(dx^{3D}_i, dy^{3D}_i, dz^{3D}_i)$, and $\theta^{3D}_i$ are the 3D center, dimension, and orientation of the $i$-th object, respectively.

\subsubsection{Challenges of 3D object detection using SAM}

Since SAM was initially trained for 2D segmentation with natural images, many inherent challenges exist to adopting SAM for LiDAR-based 3D object detection. In this section, we provide a more in-depth discussion of these challenges, which can be categorized into the following parts:

\textbf{The input data of SAM and 3D detectors are dramatically different.} 
On the one hand, the formats of input data are different. The original SAM takes 2D images as inputs, which consist of ``dense" pixels distributed evenly on the entire 2D image plane. However, for LiDAR-based 3D object detection, the inputs are LiDAR points, representing the location of ``sparse" points spread unevenly across the 3D space. On the other hand, the information contained in inputs is different. The original SAM is trained with natural images, with pixels carrying rich semantic information, while LiDAR points only carry the geometrical information of 3D scenes. To narrow the gap of input data, we use the Bird's Eye View (BEV) as the media because of its 2D format and 3D information awareness.

\textbf{The output data of SAM and 3D detectors are significantly distinct.} 
SAM outputs 2D segmentation masks indicating the possible foreground pixels, while typical 3D detectors output 3D bounding boxes. Translating the 2D segmentation masks into 3D bounding boxes is a pivotal problem. Thanks to the property of outdoor scenes that no objects stack vertically at the same position, we can leverage the BEV maps and input LiDAR points to finish the translation, and finally equipped with SAM to form a 3D detector.

\textbf{The capability of 3D perception of SAM is limited.} 
In this paper, we aim to explore zero-shot 3D object detection, which means there are no 3D samples for models to train, and since SAM is only trained with 2D images, its 3D perception capability is limited. To overcome this challenge as much as possible, we use the BEV to ``disguise" 3D information into 2D form. Moreover, because of the limited 3D capacity of SAM, its outputs will be noisy, and we propose rule-based post-processing to filter noisy masks, which helps improve performance a lot.

\subsection{Proposed Method}
\subsubsection{Overall framework}
Our method mainly contains five steps:
\begin{itemize}
    \item LiDAR-to-BEV projection translates LiDAR signals to BEV images. 
    \item  BEV post-processing modifies original BEV images with a simple operation. 
    \item SAM~\cite{kirillov2023sam} takes in modified BEV images and mesh grid prompts to segment foreground objects in BEV. In order to accelerate the segmentation process, we prune the prompts in this step without performance sacrifice.
    \item Mask post-processing filters noisy masks according to some rules drawn from priors, which reduces the number of false positives.
    \item Mask2Box finds the minimum bounding boxes of foreground masks to extract 2D boxes in BEV and then interacts with LiDAR points, predicting the final 3D bounding boxes of objects. 
\end{itemize}
In the following sections, we will describe the detailed designs for each step.

\subsubsection{LiDAR-to-BEV projection}
\label{sec:projection}
The duty of LiDAR-to-BEV projection is to translate $N_p$ LiDAR points $P = \{ (x_i, y_i, z_i) \}_{i=1}^{N_{p}}$ with range $L_x \leq x_i \leq U_x, L_y \leq y_i \leq U_y$ to a single BEV image $I \in \mathbb{R}^{H \times W \times 3}$. 

Each point will fall into a grid of BEV image, and the position of grid $(cx, cy)$ is calculated as follow:
\begin{equation}
\label{eq:bev_rank}
    cx_i = \lfloor(U_x - x_i) / s_x\rfloor,
\end{equation}
\begin{equation}
    cy_i = \lfloor(U_y - y_i) / s_y\rfloor,
\end{equation}
where $s_x, s_y$ are the pillar size of $x$ and $y$ axes, $U_x, U_y$ are the coordinate upper bounds of $x$ and $y$ axes respectively, and $\lfloor\cdot\rfloor$ means the floor function.

After obtaining the positions of grids, we need to get the values filled into the BEV image. In order to make segmentation easier, we want the BEV image to be discriminative. 

One finding is that the reflection intensity of points is useful, which means we can utilize the intensity $R = \{r_i\}_{i=1}^{N_{p}}$ to form the feature vectors of grids in a BEV image. Concretely, we first regularize the intensity to $[0, 1]$, and then take it to select color vectors from a predefined palette, which can be formally written as:
\begin{equation}
    \mathbf{c_i} = \mathrm{Palette}(\mathrm{Norm}(r_i)) \in \mathbb{R}^{3},
\end{equation}
\begin{equation}
    I[cx_i, cy_i, :] = \mathbf{c_i},
\end{equation}
where $\mathrm{Palette}: \mathbb{R} \rightarrow \mathbb{R}^3$ is the predefined palette used for translating an intensity scalar to an RGB vector.

For those grids without projected points, we simply fill in all zero vectors. Now we get a discriminative BEV image $I\in \mathbb{R}^{H \times W \times 3}$.

\subsubsection{BEV post-processing}
\label{sec: bev_postprocess}
Since SAM is trained on natural images, which contain "dense" signals and differ from the "sparse" BEV images, we need to post-process BEV images to narrow the gap. We use the morphology dilation in this paper, which can be interpreted as a max pooling, shown as Eq.~\ref{eq: dilation}.
\begin{equation}
\label{eq: dilation}
    I^{\prime} = \mathrm{MaxPool2D}(I),
\end{equation}
where $I^{\prime}$ is the BEV image after post-processing.

\subsubsection{Segmentation with SAM}
Now, we segment the BEV image using SAM, which supports various prompts like point, box, and mask prompts. Our goal in this step is to segment foreground objects as many as possible, so we choose to cover the whole image with mesh grid prompts. Specifically, we create $32 \times 32$ mesh grids evenly distributed on the image plane and regard them as point prompts to SAM. 

Although this can cover the whole image, it could be more efficient due to the natural sparsity of BEV images, with many prompts falling into empty space. Based on this observation, we prune the prompts. In particular, we project these prompts onto the BEV image, check the neighbor area of each prompt, and then discard prompts with no activated pixels around. This operation accelerates the whole pipeline dramatically, bringing 5$\times$ speed up (from 0.4 FPS to 2 FPS on a single NVIDIA GeForce RTX 4090). 

At the end of this step, we now get $N_m$ segmentation masks $M = \{ m_i \in \mathbb{R}^{H \times W} \}_{i=1}^{N_{m}}$ from SAM.

\subsubsection{Mask post-processing}
\label{sec: mask_postprocess}
Despite SAM's powerful zero-shot capability, a non-negligible domain gap still exists. Hence, the masks from SAM are noisy and need further processing. 

In scenes of autonomous driving, typical cars have certain areas and aspect ratios, which can be used to filter out some false positives in masks $M$.
In detail, we filter the noisy segmentation masks using an area threshold $[T_l^a, T_h^a]$ and an aspect ratio threshold  $[T_l^r, T_h^r]$.
With these operations, we finally obtain $N_o$ relative high-quality foreground masks $M^{\prime}=\{m_i\in \mathbb{R}^{H\times W}\}_{i=1}^{N_{o}}$, each mask corresponds to a foreground object.

\subsubsection{Mask2Box}
After the segmentation, we need to predict 3D bounding boxes $B^{3D}$ from the foreground masks $M^{\prime}$. Since BEV images already carry depth information, we can directly estimate the horizontal attributes (i.e., horizontal object center, length, width, and heading) of 3D bounding boxes from the 2D masks. Meanwhile, for the vertical attributes (i.e., vertical object center and height), LiDAR points will be utilized as extra information compensation. 

We first extract the 2D minimum bounding boxes from masks, defined as Eq.~\ref{eq:2dbbox}.
\begin{equation}
\label{eq:2dbbox}
    B^{2D} = \{ (x^{2D}_i, y^{2D}_i, dx^{2D}_i, dy^{2D}_i, \theta^{2D}_i) \}_{i=1}^{N_{o}},
\end{equation}
where $(x^{2D}_i, y^{2D}_i)$, $(dx^{2D}_i, dy^{2D}_i)$, and $\theta^{2D}_i$ are the 2D center, dimension, and rotate angle of the $i$-th object. $N_o$ is the number of objects.

 Then, we project these 2D attributes back to corresponding 3D attributes:
\begin{equation}
\label{eq:2d_to_3d_1}
    x^{3D}_i = U_x - (x^{2D}_i + 0.5) \times s_x,
\end{equation}
\begin{equation}
    y^{3D}_i = U_y - (y^{2D}_i + 0.5) \times s_y,
\end{equation}
\begin{equation}
    dx^{3D}_i = dx^{2D}_i \times s_x,
\end{equation}
\begin{equation}
    dy^{3D}_i = dy^{2D}_i \times s_y,
\end{equation}
\begin{equation}
\label{eq:2d_to_3d_2}
    \theta^{3D}_i = \theta^{2D}_i,
\end{equation}
where $U_x, U_y$ are the point cloud ranges and $s_x, s_y$ are the pillar size, defined in Sec.~\ref{sec:projection}. 

Finally, we estimate the vertical centers and heights with LiDAR points. The main idea is that we select points whose BEV projections are inside the 2D bounding boxes and calculate the vertical attributes using their vertical coordinates:
\begin{equation}
\label{eq:vertical_3d}
   Z_i = \{ z_j | (x_j, y_j, z_j) \ \mathrm{\mathbf{inside}}\ B^{3D}_i \},
\end{equation}
\begin{equation}
    dz^{3D}_i = \max{(Z_i)} - \min{(Z_i)},
\end{equation}
\begin{equation}
\label{eq:vertical_3d_2}
    z^{3D}_i = \min{(Z_i)} + \frac{dz^{3D}_i}{2}.
\end{equation}

\subsection{Experiments}
\subsubsection{Hyperparameters}
We set the point cloud range $L_x=L_y=-30.0m, U_x=U_y=30.0m$ and the pillar size $s_x = s_y = 0.1m$ by default. We use a $3 \times 3$ kernel for the dilation in BEV post-processing. For mask post-processing, we set the area thresholds $T^a_l = 200, T^a_h=5000$ pixels, and the aspect ratio thresholds $T^r_l = 1.5, T^r_h = 4$, respectively. For SAM architecture, we use the default version (ViT-H) with pre-trained weights from its official repository.

\subsubsection{Qualitative results}
\begin{figure*}[t]
\begin{center}
\includegraphics[width=0.95\linewidth]{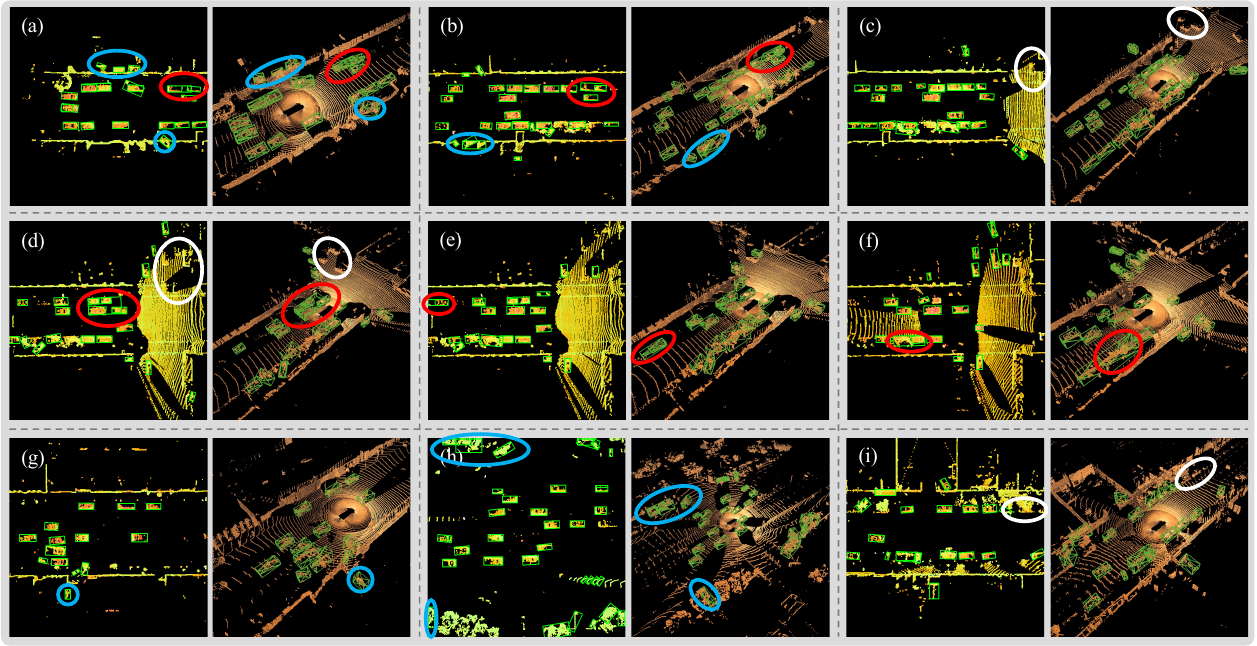}
\end{center}
    \captionof{figure}{The visualizations of results from SAM3D. Each sub-figure corresponds to a single frame. The left side of each sub-figure is the visualization of 2D bounding boxes under the Bird's Eye View (BEV), and the right is the visualization of 3D bounding boxes.}
    \label{fig:vis}
\end{figure*}

We first show the qualitative results of our method. Fig.~\ref{fig:vis} shows that relatively high-quality 2D rotated bounding boxes are generated from SAM outputs, indicating SAM's great zero-capability. It means that SAM can generate good masks without touching BEV images and 3D annotations during training. Our mask post-processing and Mask2Box module can translate foreground masks into high-quality 3D bounding boxes. It shows that for those objects that are perceived completely and have distinguishable appearances in BEV images, SAM3D can identify them easily and produce reasonable predictions.

Despite SAM's incredible power, some obvious failure cases still exist: (1) SAM will generate duplicated masks when objects are close to each other (marked as red bubbles in Fig.~\ref{fig:vis}). We argue that it is hard for SAM to identify whether these points belong to different objects or a single large object since it is not trained to handle this situation. (2) Some background objects look similar to cars in BEV images, and SAM sometimes regards them as foregrounds by mistake (marked as blue bubbles in Fig.~\ref{fig:vis}). (3) Due to truncation, occlusion, and the sparsity of LiDAR signals, some cars are partially activated in BEV images. Thus, SAM ignores these objects, leaving many false negatives (marked as white bubbles in Fig.~\ref{fig:vis}). We argue that (2) and (3) are inherent challenges for LiDAR-based 3D object detection and are even more complex for a model trained initially for 2D segmentation tasks. While our method provides some naive solutions, it still demands many efforts (e.g., more powerful background suppression or foreground completion methods can be involved) to solve these problems more perfectly. Moreover, as~\cite{ji2023samstrugglesinconcealed} depicts, SAM struggles to segment camouflaged objects. We can view those objects in (2) and (3) as camouflaged, thus getting poor results. It means we need more powerful technologies to enhance SAM's ability. 

\subsubsection{Ablation study}
We conduct ablation studies to figure out the contribution of different designs. We report the AP and APH for all experiments only for VEHICLE in the range [0, 30). 

\begin{figure*}[t]
\begin{center}
\includegraphics[width=0.9\linewidth]{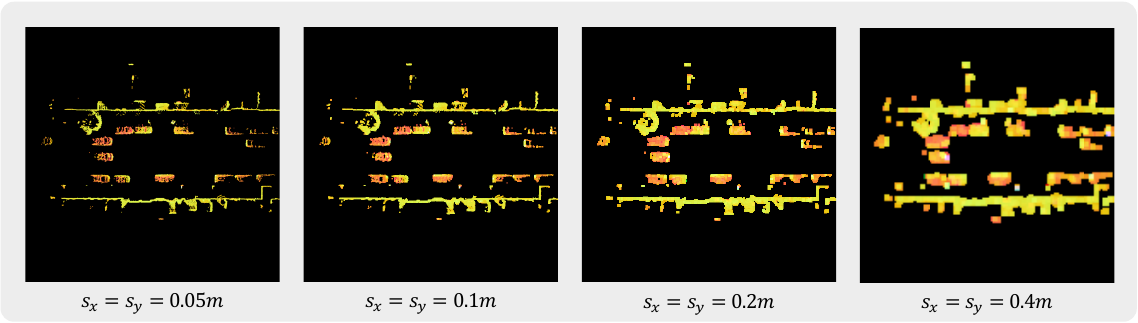}
\end{center}
      \caption{The visualization of BEV images under different pillar size settings.}
\label{fig:ablation_voxel}
\end{figure*}

\textbf{The effects of pillar size.} According to Eq.~\ref{eq:bev_rank}, the pillar size will affect the resolution of BEV images, thus influencing the segmentation results of SAM. In the letter, we report the effects of pillar size. To make its effects more intuitive, we visualize the BEV under different pillar size settings in Fig.~\ref{fig:ablation_voxel}. When using larger pillar sizes such as $0.2m$ and $0.4m$, the discretization errors are relatively large, and it is hard to distinguish different objects when they are close to each other. However, pillar sizes that are too small also harm the performance. One possible reason is that due to the high resolution of the small pillar size and the sparsity of LiDAR signals, it is difficult for individual instances to form a completely connected region. SAM tends to separate one object into many parts. We set the pillar size as $0.1m$, which is a good balance.

\begin{table}
\footnotesize
    \setlength{\tabcolsep}{2.5mm}
\centering
\caption{The results of SAM3D using different types of BEV images. We report metrics of VEHICLE in range [0,30) on Waymo \textit{validation} set.}
\label{tab:ablation_intensity}
\begin{tabular}{ ccccc }
\toprule
\multirow{2.5}{*}{BEV type} & \multicolumn{2}{c}{Level1} & \multicolumn{2}{c}{Level2} \\
\cmidrule{2-3} \cmidrule{4-5} & AP & APH & AP & APH \\
\midrule
Binary & 0.94 & 0.55 & 0.92 & 0.53 \\
Intensity & 1.93 & 1.17 & 1.88 & 1.14 \\
\rowcolor{blue!20} Intensity + Palette & \textbf{19.51} & \textbf{13.30} & \textbf{19.05} & \textbf{12.98} \\
\bottomrule
\end{tabular}
\end{table}

\textbf{The effects of reflection intensity.} The way to project points into a BEV determines the visual appearance of BEV images, affecting the segmentation. In~\ref{sec:projection}, we claim that using the reflection intensity of points and a predefined intensity-to-RGB mapping is helpful, and we evaluate its effects in this subsection. We compare our method with two other types: binary and intensity. For binary type, a pixel in a BEV image will be set to white if any point is projected into it, or it will be black otherwise. For intensity type, we use the normalized intensity as the grayscale. In Tab.~\ref{tab:ablation_intensity}, we can see that using intensity brings gains against the naive binary type, and mapping the intensity to rgb space further dramatically boosts the performance. Because the intensity and RGB mapping make BEV images more discriminative, SAM segments them more easily and precisely, thus improving the performance of the proposed method.

\begin{table}
\footnotesize
    \setlength{\tabcolsep}{1.5mm}
\centering
\caption{The ablations of post-processes. BEV post. means BEV post-processing. Area and Aspect ratio correspond to filter masks according to areas and aspect ratios in mask post-processing, respectively. We report metrics of VEHICLE in the range [0,30) on Waymo \textit{validation} set.}
\label{tab:ablation_postprocess}
\begin{tabular}{ ccc cccc }
\toprule
\multirow{2.5}{*}{BEV post.} & \multirow{2.5}{*}{Area} & \multirow{2.5}{*}{Aspect ratio} & \multicolumn{2}{c}{Level1} & \multicolumn{2}{c}{Level2} \\
\cmidrule{4-7} & & & AP & APH & AP & APH \\
\midrule
-  & \checkmark & \checkmark & 11.01 & 7.68 & 10.93 & 7.64\\
\midrule
\checkmark & - & \checkmark & 17.52 & 11.93 & 17.11 & 11.65 \\
\checkmark & \checkmark & - & 14.05 & 9.61 & 13.72 & 9.38\\
\midrule
\rowcolor{blue!20} \checkmark & \checkmark & \checkmark & \textbf{19.51} & \textbf{13.30} & \textbf{19.05} & \textbf{12.98} \\
\bottomrule
\end{tabular}
\end{table}

\textbf{The effects of BEV post-processing.} In~\ref{sec: bev_postprocess}, we use the morphology dilation to post-process the BEV image for better segmentation. We conduct experiments to figure out its effect in this subsection, shown in the first row of Tab.~\ref{tab:ablation_postprocess}. Without BEV post-processing, it drops about 8\% AP and 5\% APH on both Level 1 and Level 2. As we have claimed, BEV post-processing helps narrow the gap between BEV and natural images (training data of SAM), leading to better results.

\textbf{The effects of mask post-processing.} In Sec.~\ref{sec: mask_postprocess}, we propose the mask area and aspect ratio thresholds $(T^a_l, T^a_h), (T^r_l, T^r_h)$ to filter noisy masks from SAM. We also conduct experiments to test its effect, shown in the second and third row of Tab.~\ref{tab:ablation_postprocess}. Since a non-negligible domain gap still exists and the masks from SAM are noisy, it is obvious that all operations in mask post-processing are essential. Dropping any of them will result in a significant performance decrease compared to the full model.

\subsubsection{Comparison with fully-supervised 3D object detectors}
To better understand the gap between our method and prevailing fully-supervised 3D detectors, we list the results in Tab.~\ref{tab:compare}. Compared with traditional fully-supervised 3D detectors, our method is lagged by a significant gap. It is natural since SAM is only trained for 2D segmentation. Moreover, we observe that the difference between our methods' AP and APH is much more significant than that of others. This is because SAM is not orientation-aware, and the orientation predictions come from the minimum oriented bounding boxes estimation of segmentation masks, which is simple but noisy.

\begin{table}
\footnotesize
    \setlength{\tabcolsep}{1.5mm}
\centering
\caption{The results of SAM3D and fully-supervised 3D detectors. We report metrics of VEHICLE in range [0,30) on Waymo \textit{validation} set.}
\label{tab:compare}
\begin{tabular}{ cccccc }
\toprule
\multirow{2.5}{*}{Method} & \multirow{2.5}{*}{Training data} & \multicolumn{2}{c}{Level1} & \multicolumn{2}{c}{Level2} \\
\cmidrule{3-4} \cmidrule{5-6} & & AP & APH & AP & APH \\
\midrule
SECOND~\cite{yan2018second} & Waymo (3D Det) & 85.60 & 84.94 & 84.25 & 83.59 \\
CenterPoint~\cite{yin2021centerpoint} & Waymo (3D Det) & 81.93 & 81.20 & 80.53 & 79.82 \\
VoxelRCNN~\cite{deng2021voxelrcnn} & Waymo (3D Det) & 88.85 & 88.42 & 87.52 & 87.10 \\
PVRCNN~\cite{shi2020pvrcnn} & Waymo (3D Det) & 89.52 & 88.92 & 88.19 & 87.60 \\
PVRCNN++~\cite{shi2023pvrcnnpp} & Waymo (3D Det) & 90.08 & 89.61 & 88.77 & 88.30\\
CenterFormer~\cite{Zhou_centerformer} & Waymo (3D Det) & 91.11 & 90.58 & 89.84 & 89.31 \\
\midrule
Ours & SA-1B (2D Seg) & 19.51 & 13.30 & 19.05 & 12.98 \\
\bottomrule
\end{tabular}
\end{table}

\subsection{Discussion}
Through the qualitative results and ablation studies, we show that it is possible to leverage SAM, trained on large-scale segmentation datasets without any 3D annotation, to solve the zero-shot object detection task for outdoor scenes.

However, there are some areas for improvement in our current method, and we leave these issues to be solved in the future: 
\begin{itemize}
    \item Utilizing BEV images as representations means our method may be unsuitable for indoor scenes. Finding a better scene representation will be a good solution.
    \item Due to the occlusion, truncation, and sparsity of LiDAR points, our method generates many false negatives, especially for distant objects. Considering the information from other modalities will be helpful.
    \item Although we have already reduced the inference time by 5$\times$, the inference speed (2 FPS on a single NVIDIA GeForce RTX 4090) is still limited to the complexity of SAM, especially when the number of point prompts is large. Conducting model compression and distillation might solve this problem. 
    \item Our method currently does not support multi-class detection because of the lack of semantic label outputs from SAM. One possible solution is to leverage 3D vision-language models (e.g., CLIP Goes 3D~\cite{hegde2023clip_goes_3d}, CrowdCLIP~\cite{liang2023crowdclip}) for zero-shot classification.
\end{itemize}

We believe our method shows the great possibility and opportunity to unleash the potential of foundation models like SAM on 3D vision tasks, especially on 3D object detection. With technologies like few-shot learning and prompt engineering, we can use vision foundation models more effectively to solve 3D tasks better, especially considering the vast difference between scales of 2D and 3D data.


{\small
\bibliographystyle{ieee_fullname}
\bibliography{egbib}
}

\end{document}